%% file: neurips_2025.tex
\documentclass{article}
\PassOptionsToPackage{numbers,sort&compress}{natbib}
\usepackage[preprint]{neurips_2025}
\usepackage[utf8]{inputenc}
\usepackage[T1]{fontenc}
\usepackage{hyperref}
\usepackage{url}
\usepackage{booktabs}
\usepackage{microtype}
\usepackage{xcolor}
\usepackage{graphicx}
\usepackage{fancyvrb}
\usepackage{enumitem}
\usepackage{placeins}
\usepackage{float}
\setlist{itemsep=1pt,topsep=3pt,leftmargin=1.4em}
\hypersetup{colorlinks=true,linkcolor=[HTML]{1F4E8C},citecolor=[HTML]{1F4E8C},urlcolor=[HTML]{1F4E8C}}

\definecolor{skillA}{HTML}{CFE0F7}
\definecolor{skillB}{HTML}{FBD9C9}
\definecolor{skillS}{HTML}{E2E3E6}
\newcommand{\sA}[1]{{\setlength{\fboxsep}{0.7pt}\colorbox{skillA}{#1}}}
\newcommand{\sB}[1]{{\setlength{\fboxsep}{0.7pt}\colorbox{skillB}{#1}}}
\newcommand{\sS}[1]{{\setlength{\fboxsep}{0.7pt}\colorbox{skillS}{#1}}}
\newcommand{\bench}{\textsc{SkillClash}}

\title{One Skill Takes Over:\\ Clashing Agent Skills and a One-File Fix}

  \author{%
    Parsa Razmara\textsuperscript{1}\thanks{Corresponding author: \texttt{prazmara@usc.edu}.}\quad
    Seyedarmin Azizi\textsuperscript{1}\quad
    Ardalan Aryashad\textsuperscript{1} \\
    \textsuperscript{1}University of Southern California, Los Angeles, USA
  }

\begin{document}
\maketitle

\begin{abstract}
Agent Skills are instruction files that tell a coding agent how to do one job. Users install skills from many
authors, so two skills for the same job can be active at the same time. We ask what a model does when both skill
files are in its context: does it combine their compatible rules, or follow one skill and drop the other? We build
\bench{}, 16 pairs of same-job skills in which code checks every rule and a reference output shows the best result
that can be reached. On seven models from five families, from 4B open models to Qwen3-235B and GPT-4.1, a skill's
rules are followed 93 to 99\% of the time when it is loaded alone and 61 to 62\% when a same-job skill is loaded
next to it. Most outputs show \emph{takeover}: every rule that only one skill asks for is followed, and every rule
that only the other skill asks for is dropped. The worked examples inside the skills drive this. When each skill file
instead ends with one example that follows both skills, takeover falls to 1 to 15\% on the six models we tested.
Telling the model which skill has priority does not bring the dropped skill back, and telling it to apply both
skills brings the dropped skill back only on the two strong models and leaves the interacting rules broken. One merged
skill file written by the strongest model we used, Claude Opus 5.5, lets all seven models follow 96 to 99\% of the
achievable rules. Most of its gain over
a plain list of both skills' rules comes from its one worked example, and such an example helps only when it is
correct. Code and data: \url{https://github.com/prazmara/skillclash}.
\end{abstract}

\section{Introduction}

Agent Skills give coding agents reusable instructions for jobs such as writing commit messages, reviewing code or
documenting an API~\citep{anthropic2025skills}. A skill is a small folder with a \texttt{SKILL.md} file that holds a
short description and free-form instructions, and the agent loads it when a matching task comes in~\citep{agentskillsspec}.
In practice the instructions are a list of rules, often followed by a worked example. Skills come from many authors
and are installed side by side, and nothing stops two skills for the same job from being active together. A
developer may have a personal commit-message skill and a team one, or a vendor's code-review skill and a house one.
Some of their rules conflict, but most can be followed together. When both skills are available, does the model
combine their compatible rules, or does it follow one skill at the expense of the other?

\citet{ostm} measured this situation in practice. They installed pairs of public same-job skills in Claude Code and
found that the second skill took about one run in five away from the first. The first skill's own requirements were
met less often, while the tasks were still completed. Their fix is a hook that blocks the competing skill. In their
harness the agent opens skill files on demand, most of the loss comes from runs that never open the first skill, and
only the first skill is scored. This leaves two questions open: what does a model do when both skill files are
actually in its context, and can a fix keep the rules of both skills instead of blocking one?

\begin{figure}[t]
\centering
\begin{minipage}[t]{0.49\textwidth}
\vspace{0pt}
{\scriptsize\textbf{Skill A:} \texttt{conventional-commits}}\par\vspace{-2pt}
\begin{Verbatim}[fontsize=\tiny,commandchars=\\\{\},
  frame=single,framesep=2pt,rulecolor=\color{gray!60}]
\sA{subject starts with "feat:", "fix:", ...}  contradictory
\sA{a line "Why: ..."}                         exclusive
\sA{a line "Risk: low|medium|high"}            exclusive
\sA{a line "BREAKING CHANGE: ..."}             exclusive
\sS{a line "Tested: yes|no"}                   shared
\sA{the last line is "Signed-off-by: <name>"}  entangled
\end{Verbatim}
\end{minipage}\hfill
\begin{minipage}[t]{0.49\textwidth}
\vspace{0pt}
{\scriptsize\textbf{Skill B:} \texttt{ticket-first-commits}}\par\vspace{-2pt}
\begin{Verbatim}[fontsize=\tiny,commandchars=\\\{\},
  frame=single,framesep=2pt,rulecolor=\color{gray!60}]
\sB{subject starts with the ticket, "[PAY-311]"}  contradictory
\sB{body is a list of two or more bullets}        exclusive
\sB{a line "Component: <name>"}                   exclusive
\sB{a line "Rollback: ..."}                       exclusive
\sS{a line "Tested: yes|no"}                      shared
\sB{ends with a trailer line "Refs: <ID>"}        entangled
\end{Verbatim}
\end{minipage}

\vspace{5pt}

\begin{minipage}[t]{0.49\textwidth}
\vspace{0pt}
\centering
{\scriptsize\textbf{Both skills loaded:} skill B takes over}
\par\vspace{-2pt}
\begin{Verbatim}[fontsize=\tiny,commandchars=\\\{\},
  frame=single,framesep=2pt,rulecolor=\color{gray!60}]
\sB{[AUD-77] Add pagination to audit log API}

\sB{- Introduce limit and offset parameters to control the}
\sB{  number of records returned}
\sB{- Update API response format to include pagination}
\sB{  metadata (total, page, pages)}
\sB{Component: audit-api}
\sB{Rollback: Remove pagination parameters and revert}
\sB{  response format changes}
\sS{Tested: yes}

\sB{Refs: AUD-77}
\end{Verbatim}
\end{minipage}\hfill
\begin{minipage}[t]{0.49\textwidth}
\vspace{0pt}
\centering
{\scriptsize\textbf{One merged meta-skill:} both skills followed}
\par\vspace{-2pt}
\begin{Verbatim}[fontsize=\tiny,commandchars=\\\{\},
  frame=single,framesep=2pt,rulecolor=\color{gray!60}]
\sA{feat: add pagination to audit log API}

\sA{Why: users can now navigate through large volumes of}
\sA{  audit log entries without loading all data at once.}

\sB{- Introduce pagination parameters (page and page_size)}
\sB{  in the audit log API endpoint}
\sB{- Update the response format to include pagination}
\sB{  metadata (total, page, page_size)}

\sB{Component: audit-api}
\sA{Risk: low}
\sA{BREAKING CHANGE: none}
\sB{Rollback: revert this commit}
\sS{Tested: yes}

\sB{Refs: AUD-77}
\sA{Signed-off-by: Omar Haddad}
\end{Verbatim}
\end{minipage}

\vspace{-2pt}
\caption{A case from \bench{}. Top: two skills for the same job, with each rule shortened and labeled by its kind
(Section~\ref{sec:bench}). Bottom: outputs of Qwen3-4B for the request ``Write the commit message: add pagination to
the audit log API. Ticket AUD-77. Author: Omar Haddad. Component: audit-api.'' Lines are shaded by the skill whose rule
they follow (blue for A, orange for B, gray for the shared rule). Left: with both skills loaded, the model follows all
six rules of B and only the shared rule of A, in either load order. Right: with one meta-skill written by Claude Opus
5.5 (Section~\ref{sec:fix}), with A as the primary, the output follows all ten achievable rules.}
\label{fig:case}
\end{figure}

Figure~\ref{fig:case} shows what we find on one case. With both skills loaded, the model writes skill B's format in
full and ignores skill A, including the rules of A that conflict with nothing. We call this pattern \emph{takeover}.
With the two skills merged into one file, the same model follows the rules of both. \citet{ostm} already report that
one skill can take the place of another. We measure takeover with both files in the context and both skills scored,
so the choice of which file to open plays no role.

\paragraph{Contributions.}
\begin{enumerate}
\item \textbf{A benchmark.} \bench{} has 16 pairs of same-job skills. Code checks every rule, and for every pair a
reference output shows that the best achievable result can be reached. A co-author who had not seen our labels found
every contradictory and shared rule pair we planted (Section~\ref{sec:bench}).
\item \textbf{A diagnosis that holds across models.} On seven models from five families, from 4B open models to
GPT-4.1, a same-job skill cuts the share of a skill's rules that are followed by about a third, while an unrelated
skill changes nothing. In most outputs one skill's own rules are all followed and the other's are all dropped. The
worked examples inside the skills drive most of this. When only one skill keeps its example, that skill wins most
takeovers, and when each file ends with an example that follows both skills, takeover falls to 1 to 15\%, on the small
and the strong models alike (Section~\ref{sec:takeover}).
\item \textbf{What brings the rules back.} A priority instruction does not, on any model. An ``apply both''
instruction does on the two strong models, but the rules that interact stay broken. One merged file, a
\emph{meta-skill}, written by Claude Opus 5.5 works on all seven models. Its worked example, which follows both
skills, accounts for most of its gain over a plain list of both skills' rules, and it helps only when it is correct. A step-by-step merge that labels rule
pairs is only as good as its labels, so who writes the merge matters more than how (Section~\ref{sec:fix}).
\item \textbf{A look inside.} One direction in the activations of three small models flips which skill wins in up to
half of the takeovers, but steering along it does not make the models follow both skills (Section~\ref{sec:inside}).
\end{enumerate}

\section{Related work}

\paragraph{Agent Skills and how they fail.} Curated skills raise task pass rates, but tasks that use four or more
skills gain less, and conflicting guidance has been suggested as a cause~\citep{skillsbench}. On real software
engineering tasks, most public skills give little or no gain~\citep{han2026sweskillsbench}. A single loaded skill can
make an agent leave out or wrongly fill required parts of a task~\citep{harmfulskills}. In a large skill library, most
of the damage comes from picking a wrong skill or no skill~\citep{moreskills}. The closest work is \citet{ostm}, which
studies conflicts between co-installed same-job skills and measures, requirement by requirement, what the installed
skill loses. Our study differs in three ways: both skill files are in the context, both skills are scored against a
known best output, and the fixes we test keep both skills.

\paragraph{Conflicting instructions.} The instruction hierarchy trains a model to follow the more privileged of two
conflicting instructions~\citep{wallace2024ih}, yet models often do not enforce a stated priority, even between simple
formatting rules~\citep{geng2026controlillusion}. Our priority instruction fails in the same way. Patterns shown
earlier in a conversation can override an explicit instruction~\citep{camassa2026doasisay}, and in our setting the
worked example inside each skill plays this role. These works study one instruction against another. We study two
whole skill files of equal rank, where most of what is lost conflicts with nothing.

\paragraph{Merging and improving skills.} SkillOps merges skills whose bodies are identical~\citep{skillops}.
SkillSmith estimates from task scores whether two retrieved skills help or hurt each other, and its skills do
different jobs~\citep{skillsmith}. SkillComposer rewrites two similar skills into one more general skill and judges
the result by task pass rate~\citep{skillcomposer}. Closest to our fix, SkillFusion labels pairs of constraints in
co-loaded skills as hard conflicts, soft conflicts or redundancy, compiles the skills into one, and beats
concatenation and a one-call rewrite on task success~\citep{skillfusion}. GEPA improves prompts by reflecting on
failed runs and their feedback~\citep{gepa}, and SkillOpt edits one skill document under a validation
gate~\citep{skillopt}. On repository skills, the gains of both could not be told apart from run-to-run
noise~\citep{skillissue}. These works score task success or the merged file itself. We score every output on the rules
of both source skills against a known best output. This shows what each fix keeps and what it drops, and it shows
what makes a merge work: correct labels and one correct example that follows both skills. Our feedback loop is a
small loop of the GEPA kind, reported with intervals over jobs and with a control for how its practice requests are
chosen.

\section{\bench{}}
\label{sec:bench}

\paragraph{Jobs and skills.} A \emph{job} is one kind of text, such as a commit message, a pull request
description, a SQL query or a bug report. \bench{} has 16 jobs (Appendix~\ref{app:bench}). Each job has two skills, A
and B, with different house rules. Each skill is a \texttt{SKILL.md} file with six rules and one worked example that
follows them. We constructed the skills for this benchmark rather than collecting them from public repositories, so
that we control how their rules interact.

\paragraph{Four kinds of rules.} As in IFEval~\citep{ifeval}, a small program checks every rule. Relative to the
other skill, each rule is one of four kinds (Figure~\ref{fig:case}):
\begin{itemize}
\item \emph{contradictory}: no output can satisfy it together with its partner in the other skill, for example two
different required starts of the first line (one per skill);
\item \emph{shared}: the other skill asks for the same thing (one per skill);
\item \emph{entangled}: both rules can hold, but only if the output is arranged with care, for example two lines that
each want to come last (one per skill);
\item \emph{exclusive}: nothing in the other skill touches it (three per skill).
\end{itemize}

\paragraph{What a correct output follows.} Following both skills cannot mean following all twelve rules, because the
two contradictory rules cannot both hold. Each pair is therefore run with a stated \emph{primary} skill, whose side of
the contradiction is the correct one. This is the only place where a skill should lose, and we run every pair twice,
once with each skill as the primary. A correct output then follows ten distinct rules: all six rules of the primary,
plus the three exclusive rules and the entangled rule of the other skill. The other skill's shared rule asks for the
same thing as the primary's, so it is already counted. For every pair and each choice of primary, the benchmark
includes a reference output, such as a complete commit message, and the same code confirms that it follows all ten
rules. For the pair in Figure~\ref{fig:case}, the two reference outputs differ only in their first line. Takeover is
different from this priority: it also drops the other skill's exclusive rules, which conflict with nothing.

\paragraph{Checking the rule labels.} Because we planted the rule relations, we asked whether they are the ones a
careful reader sees. A co-author who did not write the skills and had not seen our labels read the 16 pairs and listed
every pair of rules that interact, with a label. The rater found all 16 contradictory and all 16 shared pairs with the
right label and listed no pair that we had not planted. Entangled pairs were harder to see. The rater labeled 8 of the
16 entangled pairs as entangled and 4 as contradictory, and did not list 4. In the pairs called contradictory, one rule
asks for an exact position, such as the first line, and the other only for a nearby one, such as ``at the top''; both
can hold if the lines are placed with care. The pairs not listed are length and format limits. The models also find
entangled pairs the hardest to satisfy (Section~\ref{sec:fix}).

\paragraph{What we measure.} \emph{Rules followed} is the share of a skill's six rules that an output follows.
\emph{Takeover} means that one skill's three exclusive rules are all followed and the other skill's are all dropped.
\emph{Achievable rules followed} is the share of the ten achievable rules that an output follows; an output that
follows only the primary skill gets 60\%. An output is \emph{fully correct} when it follows all ten. The last measure
is stricter: a fix can follow most achievable rules on average while few outputs follow all of them.

\paragraph{Setup.} Both skill files are placed in the system prompt, which says that every rule of every loaded skill
must be followed (Appendix~\ref{app:bench}). This removes skill selection as a source of failure. We use eight
requests per job and test seven models from five families with greedy decoding: five open models (Qwen3-4B,
Qwen2.5-7B, Mistral-7B, Llama-3.1-8B and Phi-4-mini) and two strong models called through an API (Qwen3-235B and
GPT-4.1). Qwen3-235B is the large model of the same release as Qwen3-4B. GPT-4.1 was run on four of the eight requests
per job. The 16 jobs are the unit of analysis. All intervals are 95\% bootstrap intervals over jobs, and all
comparisons are paired by job.

\section{One skill takes over}
\label{sec:takeover}

\begin{figure}[t]
\centering
\includegraphics[width=\textwidth]{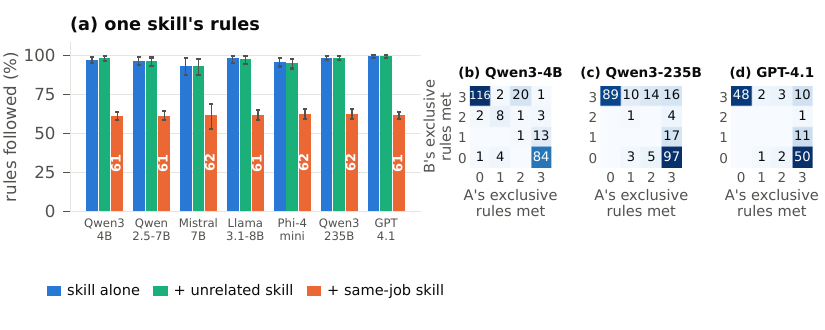}
\caption{(a) Share of a skill's rules that are followed on seven models, with 95\% intervals over the 16 jobs. (b to
d) Outputs with both skills loaded, counted by how many exclusive rules of each skill they follow. Most outputs sit in
the two corners where one skill has everything and the other has nothing, for a 4B model and for the two strong models
alike. Appendix~\ref{app:perjob} gives every model by job.}
\label{fig:takeover}
\end{figure}

\paragraph{A same-job skill hurts every model, and an unrelated skill does not.} Alone, a skill is followed almost
perfectly (Figure~\ref{fig:takeover}a). Adding a skill for a different job changes nothing, so a longer context is
not the cause. Adding a skill for the same job cuts the rules followed to 61 to 62\% on all seven models. A model with
sixty times more parameters and a strong closed model lose as much as the 4B model. One side of the contradiction has
to lose, but that explains little: on the five rules per skill that can always hold, the drop is still 28 to 35
points. The drop is about the same when both files are given in the user message instead of the system prompt (32 to
37 points on Qwen3-4B, Qwen2.5-7B, Llama-3.1-8B, Phi-4-mini, Qwen3-235B and GPT-4.1).

\paragraph{The loss is mostly all or nothing.} The outputs are rarely blends: in 45 to 78\% of outputs, one skill's
exclusive rules are all followed and the other skill's are all dropped (Figure~\ref{fig:takeover}b to d). The
exclusive rules conflict with nothing, and they are still dropped.

\begin{figure}[t]
\centering
\includegraphics[width=\textwidth]{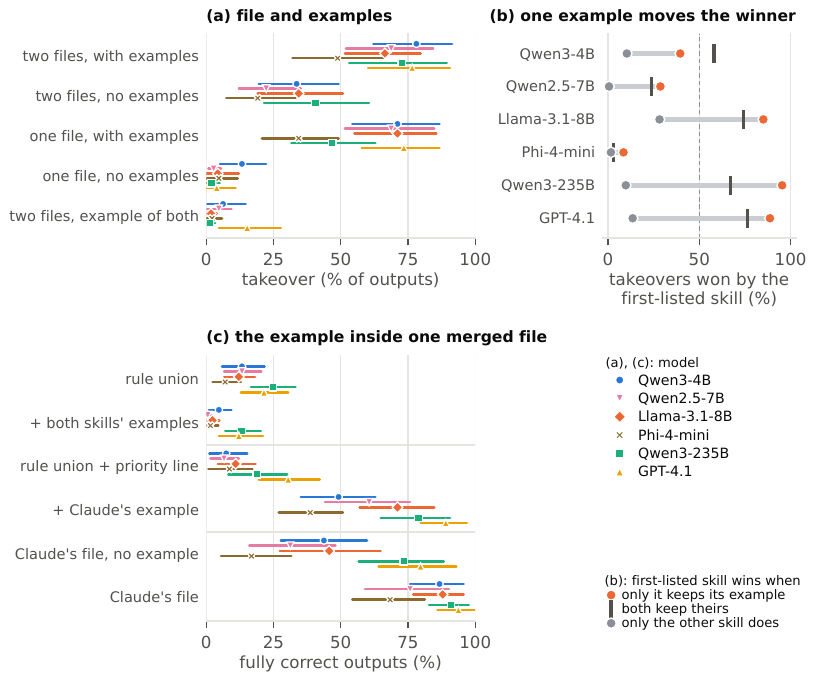}
\caption{What the worked examples do, on Qwen3-4B, Qwen2.5-7B, Llama-3.1-8B, Phi-4-mini, Qwen3-235B and GPT-4.1 (95\%
intervals over the 16 jobs). (a) Takeover with the same rules shown as two skill files or as one file, with and
without the skills' examples, and as two files that each end with the example of Claude's merged file, which follows
both skills. (b) Both skills loaded and only one keeps its example: share of the takeovers won by the first-listed
skill when only it keeps its example, when both keep theirs, and when only the other skill does. (c) One merged file,
fully correct outputs. The first two rows show the rule union without and with the two original examples, the next
two the rule union with a priority line, without and with Claude's example, and the last two Claude's file without
and with its example.}
\label{fig:trigger}
\end{figure}

\paragraph{Worked examples drive most of the takeover.} On the five open models and Qwen3-235B, takeover is 45 to
78\% with the examples and 7 to 41\% when we remove the examples from both skill files (Appendix~\ref{app:trigger}).
Three more tests on Qwen3-4B, Qwen2.5-7B, Llama-3.1-8B, Phi-4-mini, Qwen3-235B and GPT-4.1 show that the examples, and
not the two separate files, cause it (Figure~\ref{fig:trigger}). First, when both full skill files are placed in one
file, examples included, takeover is still 34 to 73\%. Second, when only one skill keeps its example, that skill wins
67 to 93\% of the takeovers over both load orders, and moving the example to the other skill moves the winner with
it. The one exception is Phi-4-mini, which follows the second-listed skill in almost every takeover. Each example
shows an output that follows one skill only, and the model writes in that format. Third, the content of the example
decides. When each of the two files ends with the same example, one that follows both skills, takeover falls to 1 to
15\% and 90 to 96\% of the achievable rules are followed, without merging the two files. On the two strong models,
this one change raises the share of fully correct outputs from 1\% to 76 and 60\%, while the ``apply both''
instruction reaches only 22 and 21\%.

Takeover does not vanish without examples. Even with no interacting rules left, four of the six models in
Appendix~\ref{app:trigger} (Qwen3-4B, Qwen2.5-7B, Phi-4-mini and Qwen3-235B) follow both skills' exclusive rules in
full 24 to 33 points less often than their single-skill runs predict, with intervals that exclude zero. So two skills
for the same job can get in each other's way even when none of their rules conflict.

\paragraph{The winner depends on the pair and the load order, not on the request.} With greedy decoding, the same
skill wins on almost every request for a fixed pair and load order (Section~\ref{sec:inside}). Swapping the load order
often changes the winner, and which position wins depends on the model. The first-listed skill wins 74 and 77\% of
the takeovers on Llama-3.1-8B and GPT-4.1, but only 24\% on Qwen2.5-7B and 3\% on Phi-4-mini. A user who does not
control the load order cannot know which skill will be followed.

\paragraph{Takeover varies by job.} Takeover is at least 75\% in 5 to 12 of the 16 jobs, depending on the model. For
SQL queries it is at most 19\% on every model. There the two rule sets constrain different parts of one statement, and
the models blend them (Appendix~\ref{app:perjob}).

\section{From two files to one}
\label{sec:fix}

We compare two families of fixes. The first keeps both skill files and adds an instruction. A \emph{priority
instruction} says that the primary wins if the skills conflict, and an \emph{``apply both'' instruction} says that the
skills are complementary and that every rule of both must be followed. The second family replaces the two files with
one file, which we call a \emph{meta-skill}. We build it in three ways.

\begin{itemize}
\item \textbf{Rule union.} Code copies the rules of both skills into one file under two headings, without the
examples, and adds a line that says which skill wins a conflict, with no model involved.
\item \textbf{Feedback loop.} A small reflective loop in the style of GEPA and SkillOpt~\citep{gepa,skillopt}
improves a meta-skill. In each of six rounds, it runs the file on three practice requests, lists the rules that each
output broke, asks the model to rewrite the file, and keeps the rewrite only if more rules are followed. Qwen3-4B both
answers the practice requests and rewrites the file.
\item \textbf{Merge by a model.} A model reads the two skill files and writes one merged skill. We tried a
\emph{plain} request to merge, and a \emph{step-by-step} merge that lists the rules, labels the pairs that interact,
writes one combined rule for each entangled pair, and copies the other rules word for word. Each of the seven models
merges for itself. An eighth merger, Claude Opus 5.5, saw only the skill files, and its files are given to all seven
models.
\end{itemize}

\begin{figure}[t]
\centering
\includegraphics[width=\textwidth]{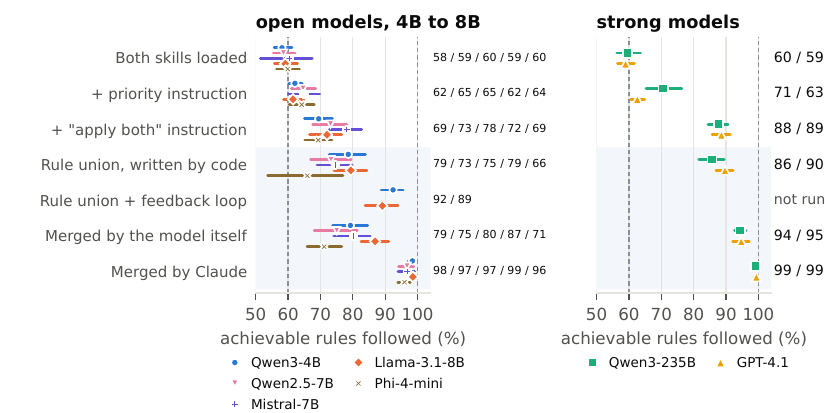}
\caption{Share of the achievable rules that are followed, with 95\% intervals over the 16 jobs. The dashed line at
60\% is what following one skill alone gives, and the line at 100\% is every achievable rule. Shaded rows use one
meta-skill file. ``Merged by the model itself'' is the plain merge, and ``Merged by Claude'' is the step-by-step merge
by Claude Opus 5.5. The feedback loop was run with Qwen3-4B, and its files were read by Qwen3-4B and Llama-3.1-8B.}
\label{fig:ladder}
\end{figure}

Figure~\ref{fig:ladder} shows the results, and Appendix~\ref{app:all} lists every condition with its interval.

\paragraph{A priority instruction does not bring the dropped skill back on any model.} With it, the achievable rules
followed stay at 62 to 71\%, close to what following one skill alone gives, and one skill still takes over in 36 to
82\% of outputs.

\paragraph{Asking for both skills works on the strong models, up to a point.} With the ``apply both'' instruction, the
open models reach 69 to 78\%, and one skill still takes over in up to 46\% of outputs. The strong models reach 88 to
89\% with no takeover, so the dropped skill comes back. The rules that interact stay broken. Both entangled rules are
followed in only 38 and 32\% of outputs, the primary keeps its side of the contradiction in 59 and 73\%, and only
about one output in five is fully correct (22 and 21\%).

\paragraph{A plain rule list in one file helps.} The rule union reaches 66 to 79\% on the open models and 86 to 90\%
on the strong ones, and takeover drops to 1 to 16\%. It needs both of its changes (Figure~\ref{fig:trigger}a). With
the examples kept, one file gains 0.6 to 9.6 points over two files, and one skill still takes over in 34 to 73\%
of outputs. Without the examples, one file gains 6 to 18 points over two files on the five open models and Qwen3-235B,
with intervals above zero on all of them except Phi-4-mini (Appendix~\ref{app:all}). On Qwen3-4B, the feedback loop
lifts the union further, to 92\%, and Llama-3.1-8B reading the same files reaches 89\%. The loop needs rule checks for
its feedback, which for real skills means a judge model or hand-written checks.

\paragraph{A file merged by the strongest model is the only fix near the ceiling on every model.} The meta-skills
written by Claude Opus 5.5 reach 96 to 99\% of the achievable rules on all seven models, with no takeover, and 68 to
94\% of outputs are fully correct. That is 11 to 29 points above the ``apply both'' instruction, and on the strong
models it gives four times as many fully correct outputs. A model that merges for itself does less well. Its plain
merge reaches 71 to 95\%, which is 5 to 25 points below Claude's file. The quality of a meta-skill follows the
strength of the model that merges, and this holds with the reader fixed: Qwen3-4B follows 79, 87, 94, 92 and 98\% of
the achievable rules when the plain merge was written by Qwen3-4B itself, Llama-3.1-8B, Qwen3-235B, GPT-4.1 and Claude
Opus 5.5, and Llama-3.1-8B as the reader shows the same pattern (Appendix~\ref{app:all}).

\paragraph{The example in the merged file does most of the work.} Each of Claude's files ends with one worked example
that follows both skills. On the four open models with this run, removing the example costs 9 to 25 points of
achievable rules; on the two strong models it costs about 2. On all six models it lowers the share of fully correct
outputs, by 14 to 52 points (Figure~\ref{fig:trigger}c). Adding the same example to the rule union written by code
lifts the union by 9 to 24 points, to within 8 points of Claude's full file on all six models. On GPT-4.1, it raises
the fully correct outputs of the rule union from 30 to 89\%. The original examples do the opposite. Added to the rule
union, they cost 0 to 14 points and bring takeover back to 32 to 55\%. Examples therefore cut both ways: an example
that follows one skill causes takeover, and an example that follows both skills is the main part of a good merge.

\begin{figure}[t]
\centering
\includegraphics[width=\textwidth]{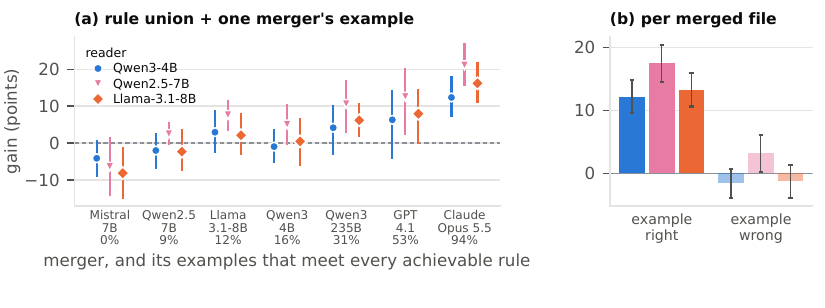}
\caption{A merged file's example helps only when it is correct. (a) Gain in achievable rules followed when one
merger's example is added to the rule union, with 95\% intervals over the 16 jobs, for files read by Qwen3-4B,
Qwen2.5-7B and Llama-3.1-8B. Mergers are ordered by the share of their examples that meet every achievable rule, given
under each name. (b) The same gain per merged file, for examples that meet every achievable rule and for examples that
do not, with the reader colors of (a).}
\label{fig:exquality}
\end{figure}

\paragraph{The example helps only when it is correct.} We added each merger's own example to the same rule union. The
gain follows how often a merger's examples meet every achievable rule. The examples of Mistral-7B, none of which does,
add nothing or cost points, and those of Claude Opus 5.5, which almost always do, add the most
(Figure~\ref{fig:exquality}). Over all merged files, a correct example adds 12 to 17 points and an incorrect one adds
little or nothing. For the same job and primary, a correct example written by another merger helps as much as Claude's
(within 1 point on each of the three readers). A strong merger helps mainly because it writes a correct example that
follows both skills far more often.

\paragraph{The step-by-step merge is only as good as its labels.} We expected the step-by-step merge to beat the
plain merge. It does so only for the strongest merger. For Claude Opus 5.5 the two are within about one point, and the
step-by-step files give more fully correct outputs on all seven models (68 to 94\% against 60 to 86\%). For the seven
models that merge for themselves, the step-by-step merge is level with their plain merge or below it
(Appendix~\ref{app:all}).

The reason is the labeling step. Claude Opus 5.5 labels all 48 interacting rule pairs correctly and adds none. The
seven models label 2 to 39 of the 48 pairs correctly, report 8 to 61 pairs that do not interact, and treat every rule
correctly in at most 14 of their 32 files. The typical errors are an entangled pair that is missed and left as two
separate rules, two independent rules that are rewritten into one, and a contradiction that is labeled as entangled.
Where a file has no such error, the step-by-step merge nearly matches the plain merge (96 against 96\% for Qwen3-235B,
and 94 against 96\% for GPT-4.1). Where it has one, it loses (84 against 94\%, and 88 against 94\%). Given the correct
labels, the step-by-step merge of Qwen3-4B gains 8 to 10 points on all seven models, and those of Qwen3-235B and
GPT-4.1 gain 4 and 3 points on their own models, which brings them to within 4 and 1 points of their plain merges.
Structure helps only when the labels are right, and only the strongest merger got them right.

Choosing the practice requests of the feedback loop so that they provoke a clash, in the spirit of curriculum
generation~\citep{agent0}, gave the same gain as plain requests when we tested it on the merged file of Qwen3-4B.

\section{Inside the model}
\label{sec:inside}

\paragraph{Can the winner be read from the activations?} In conflicts between system and user instructions, the
winning side can be read from the activations and moved by steering~\citep{balpstraffon2026choosesides}. For two
skills in one prompt we ask for more: can steering make the model follow both? We take the model's activations at the
last prompt token, before it writes anything, and train a linear probe~\citep{alain2016probes} to predict whether the
first-listed or the second-listed skill will win. For a pair and load order that the probe has seen on other requests,
there is nothing left to learn: guessing that pair's usual winner is already right for 96 to 99\% of the takeover
outputs on Qwen3-4B, Qwen2.5-7B and Llama-3.1-8B. For a pair it has not seen, we could not show on any of the three
models that the probe beats always guessing the first-listed skill (Appendix~\ref{app:steer}). With 16 held-out pairs
this test is weak, and the steering result below suggests that some general signal exists.

\begin{figure}[t]
\centering
\includegraphics[width=\textwidth]{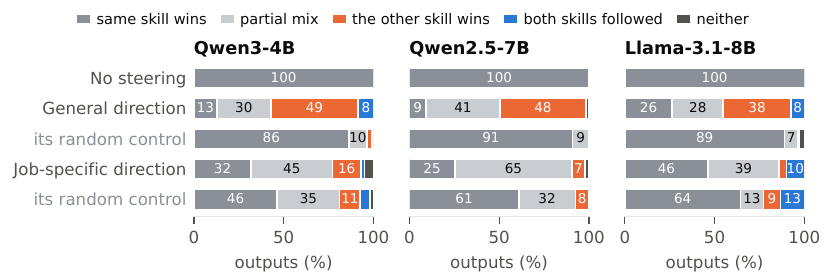}
\caption{Activation steering toward the skill that lost, on requests not used for tuning (89 to 100 outputs per model,
all of which end in a takeover without steering). Each random control is a random direction added at the same layer
and with the same strength as the row above it. Numbers inside the bars are percentages.}
\label{fig:steer}
\end{figure}

\paragraph{Can the winner be moved?} Activation steering adds a fixed vector to the output of one layer while the
model runs~\citep{turner2023steering,rimsky2024caa}. Steering at a single layer can be unstable~\citep{azizi2026sharp,saadatinia2026circuitsteer}. We steer toward the skill that lost, on requests that were not
used to choose the layer and the strength (Appendix~\ref{app:steer}). A general direction, computed from the other 15
jobs as ``first-listed skill won'' minus ``second-listed skill won'', flips the winner in 38 to 49\% of takeovers on
the three models. A random direction of the same size at the same layer flips 0 to 3\% (Figure~\ref{fig:steer}). A
direction built for the specific job flips about as many as its own random control. Steering does not make the model
follow both skills. That happens in at most 13\% of steered outputs on any of the three models, and on Llama-3.1-8B a
random direction does as well. On Qwen3-4B, stronger steering breaks the text.

So takeover behaves like a choice between two alternatives. It can be tipped the other way, but steering does not turn
it into following both skills. This matches Section~\ref{sec:fix}, where the fixes that work give the model one file,
and one example, instead of two.

\section{Discussion}

\paragraph{What this means for skill tooling.} Takeover happens after both files are loaded, so better skill
selection would not fix it, and blocking one skill~\citep{ostm} gives up that skill's rules. For the small models, the
instructions we added recover only part of the dropped skill. For the strong models an explicit instruction brings it
back, but it leaves the interacting rules broken. On every model, the best result came from merging the two skills
once, with the strongest model at hand, and loading the merged file. The merged file should end with one correct
worked example that follows both skills, and it should not copy the examples of the source skills. Where the files
cannot be merged, giving each file such an example already removes most of the takeover.

\paragraph{What is not solved.} The step-by-step merge depends on labeling the rule pairs, and only one merger labeled
them well enough. Two repairs look cheap and remain to be tested: checking each combined rule and the example against
the source rules before the file is saved, and falling back to a plain merge when the labels are in doubt. Checking a
merge on real skills also needs a judge model or hand-written checks, since real skills come without rule checkers.

\paragraph{Real skills.} In the public run summaries of \citet{ostm}, Claude Code runs that used both skills met the
requirements that only the first skill states about 8 points less often than the paired run with that skill alone
(Appendix~\ref{app:logs}). This is observational, and the effect is smaller than in our benchmark, but it points the
same way.

{\small
\bibliographystyle{plainnat}
\bibliography{refs}
}

\appendix
\raggedbottom

\FloatBarrier
\section{Benchmark details}
\label{app:bench}

\paragraph{Jobs.} Commit message, pull request description, SQL query, incident report, Python function, support
reply, changelog entry, JSON error body, code review comment, test case specification, standup update, bug report,
Dockerfile, endpoint documentation, on-call alert and user story.

\paragraph{Checks before any model is run.} A self-test confirms for every pair that each skill's example follows its
own rules, that the contradiction and the entangled pair are real on the examples, and that the reference output
for each choice of primary follows its ten rules.

\paragraph{Prompts.} The system prompt is: ``You are an assistant working in a software team's workspace. The skills
below are installed and loaded for this session. Follow every rule of every loaded skill that applies to the
request.'' The skill files follow, each inside a \texttt{<skill>} tag. The priority instruction adds: ``If two loaded
skills give conflicting instructions, the skill X takes precedence. Otherwise follow all rules of all loaded skills.''
The ``apply both'' instruction adds: ``The loaded skills are complementary, not alternatives. Apply every rule of BOTH
skills in the same output. Only where two rules cannot both hold does the rule from X win.'' The user turn is the
request plus ``Reply with the [kind of text] only, with no commentary before or after it.'' The plain merge request
is: ``Two agent skills will be installed together. Merge them into ONE skill file that replaces both. Keep every rule
of both skills. Where two rules cannot both hold, the rule from X wins. Write a complete SKILL.md with frontmatter, a
rules section and one example.''

\paragraph{Run details.} Qwen3-235B (\texttt{qwen/qwen3-235b-a22b-2507}) and GPT-4.1 (\texttt{openai/gpt-4.1}) were
called through OpenRouter at temperature 0. GPT-4.1 ran four of the eight requests per job. Qwen2.5-7B and Mistral-7B-v0.3 ran on a cloud GPU with pinned model revisions. We
replayed every answer of these runs from the saved generations with the benchmark's own prompts and checkers and found
no difference.

\paragraph{Feedback loop.} The model proposes 16 practice requests per job, of which 12 are kept. Six of them give the
feedback, and the other six pick the final file. The loop runs six rounds with three feedback requests per round.

\paragraph{Rules the model follows anyway.} A few rules are followed even with no skill loaded. If we keep only the
rules that the model follows with the skill and not without it (162 to 174 of 192), the
takeover rate is 52.1 to 84.4\% across the seven models, close to the main numbers.

\FloatBarrier
\clearpage
\section{All conditions}
\label{app:all}

\begin{table}[H]
\caption{Qwen3-4B and Llama-3.1-8B, evaluation requests (\%). Achievable: achievable rules followed, with the 95\%
interval. Ent.: both entangled rules followed. All: fully correct outputs. T.o.: takeover. The rule union of the main
text is the row with a priority line. ``Write with the primary skill, revise with the other'' is two passes. ``Rule
checklist first'' asks the model to list the rules before it answers. ``True rule kinds given'' is the step-by-step
merge with the labeling step replaced by the correct labels. Rows without a note average the two load orders.}
\label{tab:all}
\centering\scriptsize
\resizebox{\textwidth}{!}{\input{tables_v6/tab_treat}}
\end{table}

\begin{table}[H]
\caption{Strong models, with the same layout as Table~\ref{tab:all}. GPT-4.1 was run on four requests per job.}
\label{tab:allapi}
\centering\scriptsize
\resizebox{\textwidth}{!}{\input{tables_v6/tab_treat_api}}
\end{table}

\begin{table}[H]
\caption{Qwen2.5-7B, Mistral-7B and Phi-4-mini, with the same layout as Table~\ref{tab:all}.}
\label{tab:all7b}
\centering\scriptsize
\resizebox{\textwidth}{!}{\input{tables_v6/tab_treat_7b}}
\end{table}

\begin{table}[H]
\caption{The seven mergers and Claude Opus 5.5. The step-by-step columns count, out of 192 rules, 48 interacting rule
pairs and 32 files, the rules copied word for word, the pairs given the right label, the pairs reported that do not
interact, the files in which every rule is treated correctly, and the files whose example follows every achievable
rule. The right-hand columns give the achievable rules followed (\%) when the files are read by the merger itself
(for Claude Opus 5.5, the mean over the seven models), by Qwen3-4B or by Llama-3.1-8B.}
\label{tab:compiler}
\centering\scriptsize
\resizebox{\textwidth}{!}{\input{tables_v6/tab_compiler}}
\end{table}

\begin{table}[H]
\caption{Paired differences behind the claims in Section~\ref{sec:fix}: achievable rules followed, in points, as the
mean over the 16 jobs with the 95\% interval below it. Each column is the model that reads the files. --: not run.}
\label{tab:h2h}
\centering\scriptsize
\resizebox{\textwidth}{!}{
\begin{tabular}{lccccccc}
\toprule
comparison (points) & Qwen3-4B & Qwen2.5-7B & Mistral-7B & Llama-3.1-8B & Phi-4-mini & Qwen3-235B & GPT-4.1 \\
\midrule
\multicolumn{8}{l}{\emph{One file against two}} \\
One file vs.\ two files (same rules, no examples) & $+12.5$ & $+9.6$ & $+6.9$ & $+17.9$ & $+5.6$ & $+14.8$ & -- \\
 & {\tiny\color{gray}[7.8, 17.5]} & {\tiny\color{gray}[5.1, 14.3]} & {\tiny\color{gray}[0.8, 14.7]} & {\tiny\color{gray}[13.6, 22.4]} & {\tiny\color{gray}[-1.2, 11.4]} & {\tiny\color{gray}[8.6, 21.2]} & \\[1pt]
Rule union vs.\ ``apply both'' instruction & $+9.1$ & $+0.1$ & $-3.5$ & $+7.4$ & $-3.4$ & $-1.9$ & $+1.1$ \\
 & {\tiny\color{gray}[3.3, 14.9]} & {\tiny\color{gray}[-6.5, 7.0]} & {\tiny\color{gray}[-9.7, 2.3]} & {\tiny\color{gray}[2.8, 11.7]} & {\tiny\color{gray}[-14.6, 6.2]} & {\tiny\color{gray}[-6.6, 2.3]} & {\tiny\color{gray}[-2.6, 4.4]} \\[1pt]
Claude's file vs.\ ``apply both'' instruction & $+28.9$ & $+23.6$ & $+18.7$ & $+26.5$ & $+26.6$ & $+11.4$ & $+10.8$ \\
 & {\tiny\color{gray}[24.5, 33.5]} & {\tiny\color{gray}[19.0, 28.7]} & {\tiny\color{gray}[14.6, 23.5]} & {\tiny\color{gray}[22.2, 31.9]} & {\tiny\color{gray}[22.9, 30.3]} & {\tiny\color{gray}[8.8, 14.3]} & {\tiny\color{gray}[8.0, 13.8]} \\[1pt]
\midrule
\multicolumn{8}{l}{\emph{Who merges and how}} \\
Claude's file vs.\ the model's own plain merge & $+19.1$ & $+21.7$ & $+16.6$ & $+11.7$ & $+24.8$ & $+4.7$ & $+4.7$ \\
 & {\tiny\color{gray}[14.1, 24.6]} & {\tiny\color{gray}[15.7, 28.7]} & {\tiny\color{gray}[11.6, 22.7]} & {\tiny\color{gray}[8.1, 15.3]} & {\tiny\color{gray}[19.7, 29.9]} & {\tiny\color{gray}[3.0, 6.4]} & {\tiny\color{gray}[2.3, 7.3]} \\[1pt]
Step-by-step vs.\ plain merge, the model merges for itself & $-0.8$ & $+0.3$ & $-15.0$ & $-5.7$ & $-18.8$ & $-7.7$ & $-3.9$ \\
 & {\tiny\color{gray}[-6.7, 5.1]} & {\tiny\color{gray}[-6.9, 8.3]} & {\tiny\color{gray}[-23.0, -7.2]} & {\tiny\color{gray}[-10.0, -1.4]} & {\tiny\color{gray}[-29.5, -8.0]} & {\tiny\color{gray}[-14.6, -2.0]} & {\tiny\color{gray}[-13.1, 2.1]} \\[1pt]
Step-by-step vs.\ plain merge, Claude merges & $+0.7$ & $+0.4$ & $+1.8$ & $+0.5$ & $+0.5$ & $+1.3$ & $+0.8$ \\
 & {\tiny\color{gray}[-0.3, 1.7]} & {\tiny\color{gray}[-0.5, 1.3]} & {\tiny\color{gray}[0.4, 3.3]} & {\tiny\color{gray}[-0.3, 1.5]} & {\tiny\color{gray}[-0.4, 1.6]} & {\tiny\color{gray}[0.5, 2.2]} & {\tiny\color{gray}[0.1, 1.7]} \\[1pt]
\midrule
\multicolumn{8}{l}{\emph{Step-by-step merge with the correct rule kinds}} \\
Qwen3-4B's merge: correct vs.\ own labels & $+8.6$ & $+8.3$ & $+9.6$ & $+8.2$ & $+8.1$ & $+9.1$ & $+9.4$ \\
 & {\tiny\color{gray}[0.8, 16.6]} & {\tiny\color{gray}[0.6, 16.4]} & {\tiny\color{gray}[3.1, 16.4]} & {\tiny\color{gray}[0.5, 16.3]} & {\tiny\color{gray}[0.7, 15.8]} & {\tiny\color{gray}[1.8, 16.8]} & {\tiny\color{gray}[2.0, 16.8]} \\[1pt]
Qwen3-235B's merge: correct vs.\ own labels & $+4.1$ & -- & -- & $+4.4$ & -- & $+4.0$ & -- \\
 & {\tiny\color{gray}[1.4, 7.1]} & & & {\tiny\color{gray}[1.6, 7.6]} & & {\tiny\color{gray}[1.2, 7.2]} & \\[1pt]
Qwen3-235B's merge: correct labels vs.\ its plain merge & $-2.7$ & -- & -- & $-2.2$ & -- & $-3.7$ & -- \\
 & {\tiny\color{gray}[-8.0, 1.9]} & & & {\tiny\color{gray}[-7.8, 2.6]} & & {\tiny\color{gray}[-9.8, 1.4]} & \\[1pt]
GPT-4.1's merge: correct vs.\ own labels & $+3.4$ & -- & -- & $+3.5$ & -- & -- & $+3.1$ \\
 & {\tiny\color{gray}[0.2, 7.8]} & & & {\tiny\color{gray}[0.1, 7.9]} & & & {\tiny\color{gray}[-0.1, 7.5]} \\[1pt]
GPT-4.1's merge: correct labels vs.\ its plain merge & $+0.4$ & -- & -- & $+0.2$ & -- & -- & $-0.8$ \\
 & {\tiny\color{gray}[-6.3, 5.9]} & & & {\tiny\color{gray}[-5.8, 5.0]} & & & {\tiny\color{gray}[-6.6, 3.7]} \\[1pt]
\midrule
\multicolumn{8}{l}{\emph{Feedback loop (Qwen3-4B's files)}} \\
Rule union after vs.\ before the loop & $+13.8$ & -- & -- & $+9.7$ & -- & -- & -- \\
 & {\tiny\color{gray}[10.4, 17.5]} & & & {\tiny\color{gray}[5.7, 14.6]} & & & \\[1pt]
Clash-provoking vs.\ plain practice requests & $-0.2$ & -- & -- & $-0.9$ & -- & -- & -- \\
 & {\tiny\color{gray}[-2.1, 1.6]} & & & {\tiny\color{gray}[-3.8, 1.5]} & & & \\[1pt]
\bottomrule
\end{tabular}
}
\end{table}

\FloatBarrier
\section{What triggers the takeover}
\label{app:trigger}

\begin{table}[H]
\caption{Exclusive rules only (\%), for the five open models and Qwen3-235B. Alone: followed with the skill alone. Both: followed on
average with both skills loaded. Both fully: outputs that follow all exclusive rules of both skills. Expected: the
same share if each skill were followed as it is when loaded alone. Each row removes more from both skill files.}
\label{tab:trigger}
\centering\scriptsize
\resizebox{\textwidth}{!}{\input{tables_v6/tab_trigger}}
\end{table}

\FloatBarrier
\section{Takeover by job}
\label{app:perjob}

\begin{table}[H]
\caption{Both skills loaded, both load orders (\% of outputs; 16 outputs per cell, 8 for GPT-4.1).}
\label{tab:perjob}
\centering\scriptsize
\resizebox{\textwidth}{!}{\input{tables_v6/tab_perjob}}
\end{table}

\FloatBarrier
\section{Steering details}
\label{app:steer}

We add one vector to the output of one transformer block at every position, while the model reads the prompt and
while it writes. The \emph{general direction} is the mean activation at the last prompt token over prompts with both
skills loaded where the first-listed skill won, minus the mean where the second-listed skill won, taken from the
other 15 jobs. The \emph{job-specific direction} is the mean activation while the model writes with the target skill
alone, minus the same for the other skill. Each vector is scaled to a fraction $c$ of the mean activation norm of its
block. We choose the block (13, 19 or 24 of 36 on Qwen3-4B; 10, 15 or 20 of 28 on Qwen2.5-7B; 12, 17 or 22 of 32 on
Llama-3.1-8B) and $c$ on four requests per job and report the other four. We always steer toward the skill that lost
without steering, so this is a test of mechanism and not a fix. On Qwen3-4B we also tried $c=0.8$ on
the tuning requests. At that strength, up to 44\% of outputs follow no exclusive rule of either
skill.

\begin{table}[H]
\caption{Linear probe at the last prompt token: accuracy (\%) in predicting which skill wins a takeover, for requests
of a seen pair and for pairs held out. Baseline: the usual winner of that pair and order (seen pair) or always the
first-listed skill (held-out pair). The best block is picked after the fact, so its accuracy is optimistic.}
\label{tab:probe}
\centering\scriptsize
\resizebox{\textwidth}{!}{\input{tables_v6/tab_probe}}
\end{table}

\begin{table}[H]
\caption{Held-out requests whose prompts end in a takeover without steering (\% of outputs). ``Best for both'' and
``best for flipping'' are the two settings chosen on the tuning requests for each direction.}
\label{tab:steer}
\centering\scriptsize
\resizebox{\textwidth}{!}{\input{tables_v6/tab_steer}}
\end{table}

\FloatBarrier
\section{Public run summaries of One Skill Too Many}
\label{app:logs}

\begin{figure}[H]
\centering
\includegraphics[width=0.7\textwidth]{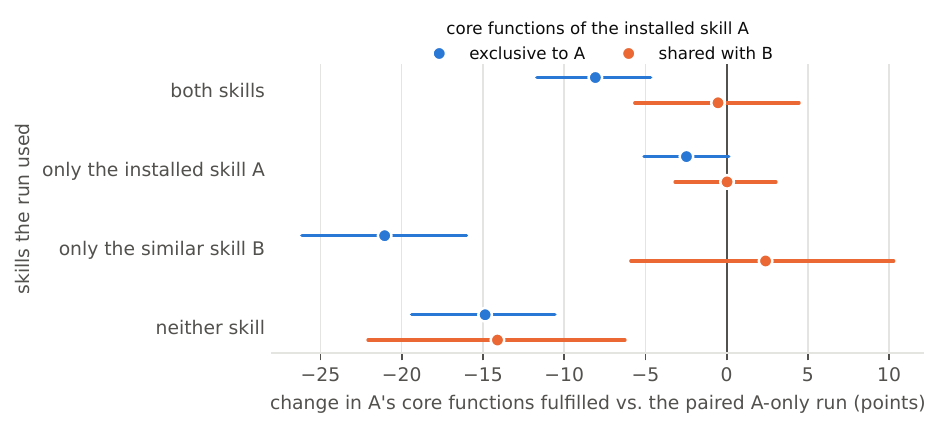}
\caption{Our reanalysis of the public run summaries of \citet{ostm} (Claude Code, real skills, both install orders):
change in the share of the installed skill A's requirements that are fulfilled, against the paired run with A alone,
by which skills the run used. ``Exclusive to A'' follows that paper's labels. Which skills a run uses is not
randomized.}
\label{fig:ostm}
\end{figure}

\FloatBarrier
\end{document}

%% file: tables_v6/tab_treat.tex
\begin{tabular}{llrrrlrrr}
\toprule
 & \multicolumn{4}{c}{model: Qwen3-4B} & \multicolumn{4}{c}{model: Llama-3.1-8B} \\
\cmidrule(lr){2-5}\cmidrule(lr){6-9}
condition & achievable & ent. & all & t.o. & achievable & ent. & all & t.o. \\
\midrule
\multicolumn{9}{l}{\emph{two skill files}} \\
Both skills loaded & 58.1 {\scriptsize $[55.7, 61.2]$} & 12 & 0 & 78 & 59.1 {\scriptsize $[55.8, 62.9]$} & 13 & 0 & 66 \\
Both skills loaded, examples removed & 64.7 {\scriptsize $[59.7, 69.6]$} & 29 & 1 & 34 & 63.1 {\scriptsize $[57.5, 69.1]$} & 30 & 3 & 34 \\
Both skills loaded, each with Claude's example & 92.9 {\scriptsize $[89.5, 96.1]$} & 90 & 60 & 6 & 94.2 {\scriptsize $[91.2, 96.6]$} & 92 & 63 & 2 \\
Write with the primary skill, revise with the other & 53.3 {\scriptsize $[50.5, 56.7]$} & 6 & 0 & 75 & 62.5 {\scriptsize $[58.3, 67.0]$} & 14 & 0 & 40 \\
Only the primary skill & 60.0 {\scriptsize $[58.9, 61.6]$} & 5 & 0 & 87 & 60.5 {\scriptsize $[59.3, 61.9]$} & 7 & 0 & 88 \\
Priority instruction & 62.1 {\scriptsize $[60.3, 64.3]$} & 10 & 0 & 74 & 61.5 {\scriptsize $[58.6, 64.9]$} & 12 & 0 & 66 \\
``Apply both'' instruction & 69.5 {\scriptsize $[65.0, 73.7]$} & 19 & 4 & 46 & 72.0 {\scriptsize $[66.6, 76.5]$} & 17 & 3 & 28 \\
``Apply both'' instruction, primary listed first & 68.2 {\scriptsize $[64.4, 72.4]$} & 17 & 3 & 60 & 74.2 {\scriptsize $[69.7, 78.3]$} & 19 & 5 & 33 \\
Rule checklist first, primary listed first & 61.7 {\scriptsize $[59.5, 64.1]$} & 10 & 0 & 79 & 74.0 {\scriptsize $[67.0, 79.8]$} & 23 & 10 & 27 \\
\midrule
\multicolumn{9}{l}{\emph{one meta-skill file}} \\
Both full skill files inside one file (no model) & 62.1 {\scriptsize $[58.1, 66.5]$} & 20 & 3 & 71 & 59.7 {\scriptsize $[56.5, 63.4]$} & 9 & 0 & 71 \\
Rule union + both original examples (no model) & 69.0 {\scriptsize $[63.9, 74.2]$} & 26 & 5 & 42 & 69.9 {\scriptsize $[66.6, 73.4]$} & 27 & 2 & 38 \\
Rule union in one file (no model) & 77.1 {\scriptsize $[70.2, 83.4]$} & 56 & 13 & 13 & 81.1 {\scriptsize $[76.2, 85.5]$} & 50 & 12 & 4 \\
Rule union + priority line (no model) & 78.6 {\scriptsize $[72.7, 83.9]$} & 55 & 7 & 16 & 79.4 {\scriptsize $[74.2, 84.3]$} & 47 & 11 & 4 \\
Plain merge, by Qwen3-4B & 79.3 {\scriptsize $[73.8, 84.6]$} & 33 & 10 & 27 & 79.7 {\scriptsize $[74.0, 85.3]$} & 37 & 11 & 28 \\
Step-by-step merge, by Qwen3-4B & 78.5 {\scriptsize $[72.3, 84.5]$} & 41 & 16 & 0 & 78.3 {\scriptsize $[71.9, 84.6]$} & 41 & 15 & 0 \\
Step-by-step merge, Qwen3-4B, true rule kinds given & 87.0 {\scriptsize $[82.3, 91.6]$} & 50 & 25 & 0 & 86.5 {\scriptsize $[81.4, 91.4]$} & 50 & 25 & 0 \\
Plain merge, by Llama-3.1-8B & 87.0 {\scriptsize $[83.2, 90.7]$} & 43 & 22 & 0 & 86.9 {\scriptsize $[82.5, 91.1]$} & 46 & 25 & 0 \\
Step-by-step merge, by Llama-3.1-8B & 80.5 {\scriptsize $[74.3, 86.0]$} & 46 & 11 & 3 & 81.2 {\scriptsize $[75.2, 86.7]$} & 45 & 14 & 3 \\
Plain merge, by Qwen3-235B & 93.9 {\scriptsize $[91.5, 96.2]$} & 66 & 51 & 0 & 93.5 {\scriptsize $[90.8, 96.1]$} & 64 & 51 & 0 \\
Step-by-step merge, by Qwen3-235B & 87.1 {\scriptsize $[80.4, 92.5]$} & 54 & 32 & 0 & 86.9 {\scriptsize $[79.7, 92.5]$} & 56 & 33 & 0 \\
Step-by-step merge, Qwen3-235B, true rule kinds given & 91.2 {\scriptsize $[85.1, 95.7]$} & 61 & 51 & 0 & 91.3 {\scriptsize $[85.3, 95.9]$} & 65 & 54 & 0 \\
Plain merge, by GPT-4.1 & 92.5 {\scriptsize $[88.8, 96.1]$} & 59 & 53 & 1 & 92.9 {\scriptsize $[89.4, 96.2]$} & 57 & 51 & 0 \\
Step-by-step merge, by GPT-4.1 & 89.5 {\scriptsize $[80.9, 95.5]$} & 59 & 45 & 0 & 89.6 {\scriptsize $[81.0, 95.5]$} & 65 & 50 & 0 \\
Step-by-step merge, GPT-4.1, true rule kinds given & 92.9 {\scriptsize $[88.2, 96.5]$} & 73 & 56 & 0 & 93.0 {\scriptsize $[88.3, 96.8]$} & 76 & 59 & 0 \\
\midrule
\multicolumn{9}{l}{\emph{one meta-skill file, after the feedback loop}} \\
Rule union + priority line, after the loop & 92.4 {\scriptsize $[88.8, 95.5]$} & 60 & 47 & 1 & 89.1 {\scriptsize $[83.8, 94.0]$} & 63 & 46 & 4 \\
Step-by-step merge, Qwen3-4B, after the loop (plain requests) & 88.4 {\scriptsize $[81.2, 94.1]$} & 67 & 43 & 0 & 87.9 {\scriptsize $[81.0, 93.6]$} & 68 & 39 & 0 \\
Step-by-step merge, Qwen3-4B, after the loop (clash requests) & 88.2 {\scriptsize $[81.6, 93.3]$} & 70 & 40 & 0 & 87.0 {\scriptsize $[80.4, 92.6]$} & 69 & 38 & 0 \\
\midrule
\multicolumn{9}{l}{\emph{one meta-skill file, merged by Claude Opus 5.5}} \\
Plain merge, by Claude Opus 5.5 & 97.7 {\scriptsize $[96.6, 98.8]$} & 90 & 79 & 0 & 98.0 {\scriptsize $[96.5, 99.3]$} & 93 & 83 & 0 \\
Step-by-step merge, by Claude Opus 5.5 & 98.4 {\scriptsize $[97.0, 99.5]$} & 98 & 87 & 0 & 98.6 {\scriptsize $[97.0, 99.6]$} & 98 & 88 & 0 \\
Step-by-step merge, by Claude Opus 5.5, example removed & 89.1 {\scriptsize $[84.1, 93.4]$} & 88 & 44 & 2 & 88.8 {\scriptsize $[83.3, 93.8]$} & 81 & 46 & 0 \\
Rule union + priority line + Claude's example & 90.9 {\scriptsize $[87.3, 94.3]$} & 80 & 49 & 6 & 95.5 {\scriptsize $[92.7, 98.0]$} & 93 & 71 & 0 \\
\bottomrule
\end{tabular}

%% file: tables_v6/tab_treat_api.tex
\begin{tabular}{llrrrlrrr}
\toprule
 & \multicolumn{4}{c}{model: Qwen3-235B} & \multicolumn{4}{c}{model: GPT-4.1} \\
\cmidrule(lr){2-5}\cmidrule(lr){6-9}
condition & achievable & ent. & all & t.o. & achievable & ent. & all & t.o. \\
\midrule
\multicolumn{9}{l}{\emph{two skill files}} \\
Both skills loaded & 59.6 {\scriptsize $[56.2, 63.6]$} & 7 & 1 & 73 & 58.9 {\scriptsize $[56.4, 61.7]$} & 7 & 1 & 77 \\
Both skills loaded, examples removed & 71.4 {\scriptsize $[65.4, 77.1]$} & 37 & 7 & 41 & -- & -- & -- & -- \\
Both skills loaded, each with Claude's example & 96.2 {\scriptsize $[93.9, 98.2]$} & 91 & 76 & 1 & 89.9 {\scriptsize $[84.7, 94.5]$} & 83 & 60 & 15 \\
Write with the primary skill, revise with the other & 59.6 {\scriptsize $[55.1, 64.5]$} & 15 & 0 & 52 & 67.3 {\scriptsize $[62.4, 72.5]$} & 19 & 0 & 38 \\
Only the primary skill & 60.7 {\scriptsize $[59.8, 61.8]$} & 5 & 0 & 89 & 60.9 {\scriptsize $[59.9, 62.3]$} & 5 & 0 & 92 \\
Priority instruction & 70.6 {\scriptsize $[65.3, 76.3]$} & 19 & 9 & 56 & 62.6 {\scriptsize $[60.5, 64.8]$} & 11 & 0 & 82 \\
``Apply both'' instruction & 87.6 {\scriptsize $[84.4, 90.6]$} & 38 & 22 & 0 & 88.6 {\scriptsize $[85.7, 91.3]$} & 32 & 21 & 0 \\
``Apply both'' instruction, primary listed first & 88.4 {\scriptsize $[85.0, 91.6]$} & 35 & 21 & 0 & 89.1 {\scriptsize $[86.0, 92.0]$} & 31 & 20 & 1 \\
Rule checklist first, primary listed first & 71.1 {\scriptsize $[65.5, 77.5]$} & 26 & 14 & 60 & 63.4 {\scriptsize $[60.5, 67.1]$} & 12 & 1 & 84 \\
\midrule
\multicolumn{9}{l}{\emph{one meta-skill file}} \\
Both full skill files inside one file (no model) & 69.1 {\scriptsize $[64.0, 74.5]$} & 19 & 6 & 47 & 60.5 {\scriptsize $[57.6, 63.9]$} & 12 & 1 & 73 \\
Rule union + both original examples (no model) & 76.6 {\scriptsize $[70.5, 82.6]$} & 40 & 13 & 32 & 72.7 {\scriptsize $[65.2, 80.3]$} & 42 & 12 & 49 \\
Rule union in one file (no model) & 86.2 {\scriptsize $[81.1, 90.2]$} & 69 & 25 & 2 & 86.6 {\scriptsize $[83.0, 89.8]$} & 60 & 21 & 4 \\
Rule union + priority line (no model) & 85.7 {\scriptsize $[81.6, 89.4]$} & 72 & 19 & 1 & 89.7 {\scriptsize $[86.9, 92.2]$} & 68 & 30 & 5 \\
Plain merge, by Qwen3-4B & 80.8 {\scriptsize $[75.2, 86.2]$} & 34 & 13 & 24 & 81.6 {\scriptsize $[75.6, 87.3]$} & 38 & 16 & 27 \\
Step-by-step merge, by Qwen3-4B & 78.4 {\scriptsize $[72.2, 84.4]$} & 43 & 17 & 0 & 80.0 {\scriptsize $[73.8, 85.9]$} & 44 & 15 & 0 \\
Step-by-step merge, Qwen3-4B, true rule kinds given & 87.4 {\scriptsize $[82.7, 91.9]$} & 48 & 29 & 0 & 89.4 {\scriptsize $[85.7, 92.9]$} & 49 & 30 & 0 \\
Plain merge, by Qwen3-235B & 94.3 {\scriptsize $[92.6, 96.1]$} & 64 & 51 & 0 & -- & -- & -- & -- \\
Step-by-step merge, by Qwen3-235B & 86.6 {\scriptsize $[79.5, 92.3]$} & 54 & 31 & 0 & -- & -- & -- & -- \\
Step-by-step merge, Qwen3-235B, true rule kinds given & 90.7 {\scriptsize $[84.4, 95.5]$} & 59 & 48 & 0 & -- & -- & -- & -- \\
Plain merge, by GPT-4.1 & -- & -- & -- & -- & 94.7 {\scriptsize $[92.0, 97.2]$} & 59 & 57 & 0 \\
Step-by-step merge, by GPT-4.1 & -- & -- & -- & -- & 90.8 {\scriptsize $[82.0, 96.6]$} & 62 & 53 & 0 \\
Step-by-step merge, GPT-4.1, true rule kinds given & -- & -- & -- & -- & 93.9 {\scriptsize $[89.1, 97.6]$} & 76 & 63 & 0 \\
\midrule
\multicolumn{9}{l}{\emph{one meta-skill file, after the feedback loop}} \\
Step-by-step merge, Qwen3-4B, after the loop (clash requests) & 88.0 {\scriptsize $[81.5, 93.0]$} & 69 & 39 & 0 & 88.8 {\scriptsize $[82.2, 93.8]$} & 72 & 41 & 0 \\
\midrule
\multicolumn{9}{l}{\emph{one meta-skill file, merged by Claude Opus 5.5}} \\
Plain merge, by Claude Opus 5.5 & 97.7 {\scriptsize $[96.3, 98.9]$} & 91 & 80 & 0 & 98.6 {\scriptsize $[97.6, 99.5]$} & 92 & 86 & 0 \\
Step-by-step merge, by Claude Opus 5.5 & 99.0 {\scriptsize $[98.1, 99.8]$} & 99 & 91 & 0 & 99.4 {\scriptsize $[98.6, 100.0]$} & 98 & 94 & 0 \\
Step-by-step merge, by Claude Opus 5.5, example removed & 96.7 {\scriptsize $[94.4, 98.7]$} & 98 & 73 & 0 & 97.7 {\scriptsize $[95.6, 99.3]$} & 97 & 80 & 0 \\
Rule union + priority line + Claude's example & 97.5 {\scriptsize $[95.5, 99.1]$} & 97 & 79 & 0 & 98.9 {\scriptsize $[98.0, 99.7]$} & 97 & 89 & 0 \\
\bottomrule
\end{tabular}

%% file: tables_v6/tab_treat_7b.tex
\begin{tabular}{llrrrlrrrlrrr}
\toprule
 & \multicolumn{4}{c}{model: Qwen2.5-7B} & \multicolumn{4}{c}{model: Mistral-7B} & \multicolumn{4}{c}{model: Phi-4-mini} \\
\cmidrule(lr){2-5}\cmidrule(lr){6-9}\cmidrule(lr){10-13}
condition & achievable & ent. & all & t.o. & achievable & ent. & all & t.o. & achievable & ent. & all & t.o. \\
\midrule
\multicolumn{13}{l}{\emph{two skill files}} \\
Both skills loaded & 58.6 {\scriptsize $[55.4, 62.2]$} & 8 & 0 & 69 & 60.3 {\scriptsize $[51.4, 67.4]$} & 16 & 2 & 45 & 59.9 {\scriptsize $[56.4, 63.5]$} & 19 & 0 & 49 \\
Both skills loaded, examples removed & 65.7 {\scriptsize $[60.1, 71.3]$} & 48 & 3 & 22 & 68.4 {\scriptsize $[60.6, 75.4]$} & 43 & 5 & 7 & 59.3 {\scriptsize $[49.3, 68.1]$} & 43 & 3 & 19 \\
Both skills loaded, each with Claude's example & 90.7 {\scriptsize $[87.5, 93.6]$} & 87 & 49 & 5 & -- & -- & -- & -- & 91.0 {\scriptsize $[87.6, 94.1]$} & 81 & 50 & 2 \\
Write with the primary skill, revise with the other & 59.2 {\scriptsize $[54.8, 64.5]$} & 16 & 0 & 56 & 60.4 {\scriptsize $[56.0, 65.6]$} & 19 & 0 & 46 & 56.5 {\scriptsize $[53.4, 59.8]$} & 15 & 0 & 62 \\
Only the primary skill & 59.8 {\scriptsize $[58.5, 61.2]$} & 4 & 0 & 85 & 57.8 {\scriptsize $[53.9, 60.8]$} & 5 & 0 & 82 & 59.7 {\scriptsize $[58.4, 61.1]$} & 6 & 0 & 82 \\
Priority instruction & 64.6 {\scriptsize $[61.0, 68.5]$} & 11 & 0 & 55 & 64.9 {\scriptsize $[60.1, 69.8]$} & 15 & 1 & 48 & 64.2 {\scriptsize $[60.3, 68.1]$} & 20 & 1 & 36 \\
``Apply both'' instruction & 73.1 {\scriptsize $[67.6, 78.2]$} & 21 & 3 & 21 & 78.1 {\scriptsize $[72.8, 82.7]$} & 23 & 7 & 8 & 69.3 {\scriptsize $[65.1, 73.6]$} & 23 & 2 & 21 \\
``Apply both'' instruction, primary listed first & 73.3 {\scriptsize $[66.8, 78.8]$} & 18 & 3 & 20 & 80.1 {\scriptsize $[74.2, 85.4]$} & 25 & 10 & 10 & 69.0 {\scriptsize $[64.6, 73.7]$} & 24 & 3 & 13 \\
Rule checklist first, primary listed first & 74.1 {\scriptsize $[70.3, 77.9]$} & 22 & 4 & 20 & 69.8 {\scriptsize $[66.5, 73.0]$} & 18 & 2 & 29 & 66.8 {\scriptsize $[62.5, 71.2]$} & 16 & 1 & 18 \\
\midrule
\multicolumn{13}{l}{\emph{one meta-skill file}} \\
Both full skill files inside one file (no model) & 60.3 {\scriptsize $[56.5, 64.7]$} & 14 & 1 & 69 & -- & -- & -- & -- & 61.2 {\scriptsize $[57.0, 65.4]$} & 23 & 0 & 34 \\
Rule union + both original examples (no model) & 62.4 {\scriptsize $[59.3, 65.8]$} & 15 & 1 & 55 & -- & -- & -- & -- & 65.1 {\scriptsize $[60.5, 70.2]$} & 20 & 2 & 34 \\
Rule union in one file (no model) & 75.3 {\scriptsize $[68.5, 81.6]$} & 50 & 13 & 3 & 75.3 {\scriptsize $[69.0, 81.1]$} & 43 & 8 & 4 & 64.9 {\scriptsize $[51.7, 76.3]$} & 38 & 7 & 5 \\
Rule union + priority line (no model) & 73.2 {\scriptsize $[67.0, 79.5]$} & 50 & 7 & 3 & 74.6 {\scriptsize $[68.9, 79.8]$} & 46 & 7 & 4 & 65.9 {\scriptsize $[53.7, 77.0]$} & 43 & 9 & 4 \\
Plain merge, by Qwen3-4B & 78.6 {\scriptsize $[72.3, 84.6]$} & 32 & 10 & 28 & 76.6 {\scriptsize $[71.0, 82.2]$} & 27 & 10 & 25 & 77.2 {\scriptsize $[71.3, 83.0]$} & 34 & 9 & 25 \\
Step-by-step merge, by Qwen3-4B & 77.8 {\scriptsize $[71.7, 83.9]$} & 41 & 14 & 0 & 74.5 {\scriptsize $[68.3, 80.6]$} & 35 & 11 & 0 & 76.8 {\scriptsize $[70.7, 83.2]$} & 40 & 15 & 0 \\
Step-by-step merge, Qwen3-4B, true rule kinds given & 86.1 {\scriptsize $[80.9, 91.2]$} & 48 & 23 & 0 & 84.1 {\scriptsize $[77.8, 90.0]$} & 50 & 23 & 0 & 85.0 {\scriptsize $[79.5, 90.2]$} & 50 & 25 & 1 \\
\midrule
\multicolumn{13}{l}{\emph{one meta-skill file, after the feedback loop}} \\
Step-by-step merge, Qwen3-4B, after the loop (clash requests) & 86.6 {\scriptsize $[80.0, 92.2]$} & 68 & 35 & 0 & 83.6 {\scriptsize $[77.3, 89.2]$} & 61 & 29 & 0 & 85.3 {\scriptsize $[78.9, 90.7]$} & 66 & 30 & 0 \\
\midrule
\multicolumn{13}{l}{\emph{one meta-skill file, merged by Claude Opus 5.5}} \\
Plain merge, by Claude Opus 5.5 & 96.4 {\scriptsize $[94.1, 98.3]$} & 90 & 68 & 0 & 95.0 {\scriptsize $[92.5, 97.3]$} & 84 & 60 & 0 & 95.4 {\scriptsize $[93.0, 97.6]$} & 85 & 65 & 0 \\
Step-by-step merge, by Claude Opus 5.5 & 96.7 {\scriptsize $[94.1, 98.8]$} & 96 & 76 & 0 & 96.8 {\scriptsize $[93.9, 99.1]$} & 91 & 77 & 0 & 95.9 {\scriptsize $[93.9, 97.7]$} & 95 & 68 & 0 \\
Step-by-step merge, by Claude Opus 5.5, example removed & 83.2 {\scriptsize $[76.8, 89.3]$} & 85 & 31 & 1 & -- & -- & -- & -- & 71.2 {\scriptsize $[60.7, 80.8]$} & 68 & 17 & 4 \\
Rule union + priority line + Claude's example & 94.3 {\scriptsize $[91.2, 97.0]$} & 90 & 61 & 0 & -- & -- & -- & -- & 90.0 {\scriptsize $[87.2, 92.8]$} & 84 & 39 & 0 \\
\bottomrule
\end{tabular}

%% file: tables_v6/tab_compiler.tex
\begin{tabular}{lrrrrrrrrrrr}
\toprule
 & \multicolumn{5}{c}{step-by-step merge} & \multicolumn{2}{c}{files read by the merger itself} & \multicolumn{2}{c}{files read by Qwen3-4B} & \multicolumn{2}{c}{files read by Llama-3.1-8B} \\
\cmidrule(lr){2-6}\cmidrule(lr){7-8}\cmidrule(lr){9-10}\cmidrule(lr){11-12}
merger & rules copied & right labels & false pairs & clean files & good examples & step-by-step & plain & step-by-step & plain & step-by-step & plain \\
\midrule
Qwen3-4B & 190/192 & 24/48 & 38 & 2/32 & 5/32 & 78.5 & 79.3 & 78.5 & 79.3 & 78.3 & 79.7 \\
Qwen2.5-7B & 189/192 & 8/48 & 8 & 0/32 & 3/32 & 75.4 & 75.0 & -- & -- & -- & -- \\
Mistral-7B & 117/192 & 4/48 & 61 & 0/32 & 0/32 & 65.2 & 80.2 & -- & -- & -- & -- \\
Llama-3.1-8B & 182/192 & 8/48 & 33 & 0/32 & 4/32 & 81.2 & 86.9 & 80.5 & 87.0 & 81.2 & 86.9 \\
Phi-4-mini & 120/192 & 2/48 & 30 & 0/32 & 2/32 & 52.4 & 71.1 & -- & -- & -- & -- \\
Qwen3-235B & 188/192 & 32/48 & 18 & 6/32 & 10/32 & 86.6 & 94.3 & 87.1 & 93.9 & 86.9 & 93.5 \\
GPT-4.1 & 192/192 & 39/48 & 8 & 14/32 & 17/32 & 90.8 & 94.7 & 89.5 & 92.5 & 89.6 & 92.9 \\
Claude Opus 5.5 & 192/192 & 48/48 & 0 & 32/32 & 30/32 & 97.8 & 97.0 & 98.4 & 97.7 & 98.6 & 98.0 \\
\bottomrule
\end{tabular}

%% file: tables_v6/tab_trigger.tex
\begin{tabular}{llrrlrrl}
\toprule
model & skills shown & alone & both & takeover & both fully & expected & both fully $-$ expected \\
\midrule
Qwen3-4B & full skills, with examples & 96.9 & 53.4 & 78.1 {\scriptsize $[62.1, 91.4]$} & 0.4 & 86.7 & $-86.3$ {\scriptsize $[-96.9, -73.0]$} \\
 & examples removed & 92.8 & 63.5 & 33.6 {\scriptsize $[19.5, 49.2]$} & 16.0 & 62.5 & $-46.5$ {\scriptsize $[-69.5, -21.5]$} \\
 & + contradictory rules removed & 93.8 & 68.8 & 25.4 {\scriptsize $[13.3, 39.8]$} & 26.2 & 68.8 & $-42.6$ {\scriptsize $[-59.4, -25.4]$} \\
 & + entangled rules removed (fully compatible) & 92.4 & 70.8 & 23.0 {\scriptsize $[9.4, 39.1]$} & 30.5 & 63.3 & $-32.8$ {\scriptsize $[-50.8, -16.4]$} \\
\midrule
Qwen2.5-7B & full skills, with examples & 95.8 & 54.6 & 68.8 {\scriptsize $[52.0, 84.4]$} & 5.9 & 78.9 & $-73.0$ {\scriptsize $[-88.7, -55.5]$} \\
 & examples removed & 87.0 & 62.2 & 22.3 {\scriptsize $[12.1, 35.2]$} & 14.1 & 57.0 & $-43.0$ {\scriptsize $[-60.5, -25.0]$} \\
 & + contradictory rules removed & 89.7 & 72.0 & 15.2 {\scriptsize $[3.9, 30.1]$} & 24.2 & 65.6 & $-41.4$ {\scriptsize $[-58.2, -25.0]$} \\
 & + entangled rules removed (fully compatible) & 88.2 & 73.4 & 6.2 {\scriptsize $[0.0, 17.6]$} & 34.0 & 57.8 & $-23.8$ {\scriptsize $[-39.8, -8.6]$} \\
\midrule
Mistral-7B & full skills, with examples & 92.4 & 59.4 & 45.3 {\scriptsize $[27.7, 63.7]$} & 18.0 & 74.2 & $-56.2$ {\scriptsize $[-74.6, -36.7]$} \\
 & examples removed & 81.4 & 69.1 & 6.6 {\scriptsize $[0.8, 14.8]$} & 25.0 & 42.2 & $-17.2$ {\scriptsize $[-28.9, -6.6]$} \\
 & + contradictory rules removed & 78.6 & 72.6 & 1.6 {\scriptsize $[0.0, 4.7]$} & 34.8 & 34.4 & $+0.4$ {\scriptsize $[-20.7, 19.5]$} \\
 & + entangled rules removed (fully compatible) & 78.6 & 70.8 & 4.3 {\scriptsize $[0.4, 10.9]$} & 32.0 & 28.1 & $+3.9$ {\scriptsize $[-11.7, 17.6]$} \\
\midrule
Llama-3.1-8B & full skills, with examples & 97.3 & 55.1 & 66.4 {\scriptsize $[51.6, 79.7]$} & 3.9 & 89.1 & $-85.2$ {\scriptsize $[-95.7, -71.5]$} \\
 & examples removed & 94.0 & 63.8 & 34.4 {\scriptsize $[19.1, 50.8]$} & 19.5 & 71.1 & $-51.6$ {\scriptsize $[-69.1, -32.0]$} \\
 & + contradictory rules removed & 86.5 & 67.8 & 28.5 {\scriptsize $[14.1, 45.3]$} & 31.6 & 61.7 & $-30.1$ {\scriptsize $[-54.3, -4.3]$} \\
 & + entangled rules removed (fully compatible) & 82.7 & 68.2 & 26.6 {\scriptsize $[10.9, 44.9]$} & 29.7 & 48.4 & $-18.8$ {\scriptsize $[-41.8, 5.5]$} \\
\midrule
Phi-4-mini & full skills, with examples & 95.2 & 55.3 & 48.8 {\scriptsize $[32.0, 66.4]$} & 1.2 & 76.6 & $-75.4$ {\scriptsize $[-90.6, -57.8]$} \\
 & examples removed & 70.4 & 56.6 & 19.1 {\scriptsize $[7.4, 33.2]$} & 12.1 & 37.5 & $-25.4$ {\scriptsize $[-41.4, -10.5]$} \\
 & + contradictory rules removed & 81.1 & 56.7 & 17.2 {\scriptsize $[6.2, 31.2]$} & 10.2 & 50.0 & $-39.8$ {\scriptsize $[-59.4, -20.7]$} \\
 & + entangled rules removed (fully compatible) & 80.1 & 60.7 & 9.8 {\scriptsize $[2.0, 21.9]$} & 20.3 & 45.3 & $-25.0$ {\scriptsize $[-39.1, -11.7]$} \\
\midrule
Qwen3-235B & full skills, with examples & 98.3 & 56.5 & 72.7 {\scriptsize $[53.1, 89.5]$} & 6.2 & 90.6 & $-84.4$ {\scriptsize $[-95.7, -71.1]$} \\
 & examples removed & 96.1 & 71.3 & 40.6 {\scriptsize $[21.5, 60.5]$} & 30.5 & 82.0 & $-51.6$ {\scriptsize $[-70.7, -32.4]$} \\
 & + contradictory rules removed & 96.5 & 75.5 & 27.7 {\scriptsize $[12.5, 44.9]$} & 41.4 & 83.6 & $-42.2$ {\scriptsize $[-62.9, -22.7]$} \\
 & + entangled rules removed (fully compatible) & 96.1 & 79.3 & 20.3 {\scriptsize $[6.2, 37.5]$} & 47.7 & 80.5 & $-32.8$ {\scriptsize $[-51.2, -15.6]$} \\
\bottomrule
\end{tabular}

%% file: tables_v6/tab_perjob.tex
\begin{tabular}{llrrrrrrrrrrrrrr}
\toprule
 & & \multicolumn{2}{c}{Qwen3-4B} & \multicolumn{2}{c}{Qwen2.5-7B} & \multicolumn{2}{c}{Mistral-7B} & \multicolumn{2}{c}{Llama-3.1-8B} & \multicolumn{2}{c}{Phi-4-mini} & \multicolumn{2}{c}{Qwen3-235B} & \multicolumn{2}{c}{GPT-4.1} \\
job & artifact & takeover & both fully & takeover & both fully & takeover & both fully & takeover & both fully & takeover & both fully & takeover & both fully & takeover & both fully \\
\midrule
commit & commit message & 100 & 0 & 88 & 0 & 50 & 31 & 100 & 0 & 19 & 0 & 100 & 0 & 62 & 38 \\
pr & pull request description & 81 & 0 & 94 & 0 & 12 & 44 & 50 & 31 & 38 & 0 & 100 & 0 & 75 & 25 \\
sql & SQL query & 0 & 0 & 6 & 25 & 0 & 12 & 0 & 19 & 19 & 6 & 0 & 12 & 12 & 12 \\
incident & incident report & 38 & 6 & 12 & 0 & 50 & 0 & 50 & 0 & 0 & 0 & 12 & 0 & 0 & 0 \\
python & Python function & 31 & 0 & 25 & 0 & 12 & 0 & 12 & 0 & 19 & 0 & 31 & 6 & 62 & 0 \\
support & support reply & 50 & 0 & 62 & 0 & 6 & 0 & 62 & 0 & 31 & 0 & 38 & 0 & 88 & 0 \\
changelog & changelog entry & 100 & 0 & 100 & 0 & 94 & 0 & 69 & 0 & 50 & 0 & 100 & 0 & 100 & 0 \\
apierror & JSON error response body & 94 & 0 & 100 & 0 & 0 & 0 & 81 & 0 & 0 & 6 & 100 & 0 & 100 & 0 \\
review & code review comment & 94 & 0 & 19 & 50 & 81 & 6 & 94 & 0 & 50 & 6 & 100 & 0 & 100 & 0 \\
testcase & test case specification & 88 & 0 & 100 & 0 & 6 & 88 & 56 & 6 & 94 & 0 & 88 & 0 & 100 & 0 \\
standup & daily standup update & 100 & 0 & 100 & 0 & 100 & 0 & 100 & 0 & 100 & 0 & 100 & 0 & 100 & 0 \\
bugreport & bug report & 100 & 0 & 75 & 0 & 56 & 31 & 94 & 0 & 100 & 0 & 88 & 12 & 100 & 0 \\
dockerfile & Dockerfile & 81 & 0 & 56 & 12 & 94 & 0 & 81 & 6 & 94 & 0 & 12 & 62 & 50 & 25 \\
apidoc & REST endpoint documentation entry & 100 & 0 & 100 & 0 & 31 & 6 & 62 & 0 & 94 & 0 & 100 & 0 & 100 & 0 \\
alert & on-call alert notification & 100 & 0 & 88 & 6 & 31 & 69 & 100 & 0 & 25 & 0 & 100 & 0 & 100 & 0 \\
userstory & user story ticket & 94 & 0 & 75 & 0 & 100 & 0 & 50 & 0 & 50 & 0 & 94 & 6 & 75 & 25 \\
\bottomrule
\end{tabular}

%% file: tables_v6/tab_probe.tex
\begin{tabular}{llrrrlr}
\toprule
model & held out & baseline & text features & probe & probe $-$ baseline & best block, post hoc \\
\midrule
Qwen3-4B & new skill pair & 58.0 & 50.0 & 67.0 & $+6.2$ {\scriptsize $[-15.0, 24.4]$} & 74.0 (26) \\
Qwen3-4B & new request & 96.5 & 58.0 & 96.0 & $-1.3$ {\scriptsize $[-4.0, 0.0]$} & 96.5 (22) \\
Qwen2.5-7B & new skill pair & 76.1 & 75.6 & 76.1 & $-2.2$ {\scriptsize $[-15.5, 9.7]$} & 88.6 (22) \\
Qwen2.5-7B & new request & 98.9 & 79.0 & 98.3 & $-1.6$ {\scriptsize $[-4.7, 0.0]$} & 98.3 (18) \\
Llama-3.1-8B & new skill pair & 74.1 & 67.1 & 70.6 & $-5.0$ {\scriptsize $[-15.4, 0.9]$} & 79.4 (16) \\
Llama-3.1-8B & new request & 97.1 & 72.9 & 96.5 & $-0.8$ {\scriptsize $[-2.5, 0.0]$} & 97.1 (15) \\
\bottomrule
\end{tabular}

%% file: tables_v6/tab_steer.tex
\begin{tabular}{llrlrrrr}
\toprule
model & steering & stays & flips & both & partial & neither & achievable \\
\midrule
Qwen3-4B & no steering & 100 & 0.0 {\scriptsize $[0.0, 0.0]$} & 0 & 0 & 0 & 56.8 \\
Qwen3-4B & job-specific direction, best for both (block 13, $c$=0.4) & 46 & 14.6 {\scriptsize $[5.0, 25.2]$} & 5 & 34 & 0 & 60.0 \\
Qwen3-4B & job-specific direction, best for flipping (block 13, $c$=0.6) & 32 & 15.9 {\scriptsize $[3.3, 32.0]$} & 2 & 45 & 5 & 51.2 \\
Qwen3-4B & random control for the row above (block 13, $c$=0.6) & 46 & 11.4 {\scriptsize $[3.3, 21.4]$} & 6 & 35 & 2 & 59.3 \\
Qwen3-4B & general direction, best for both (block 19, $c$=0.4) & 35 & 23.3 {\scriptsize $[8.8, 38.9]$} & 13 & 28 & 1 & 62.4 \\
Qwen3-4B & general direction, best for flipping (block 24, $c$=0.4) & 13 & 48.8 {\scriptsize $[29.6, 68.3]$} & 8 & 30 & 0 & 59.2 \\
Qwen3-4B & random control for the row above (block 24, $c$=0.4) & 86 & 2.9 {\scriptsize $[0.0, 6.1]$} & 1 & 10 & 0 & 56.0 \\
Qwen2.5-7B & no steering & 100 & 0.0 {\scriptsize $[0.0, 0.0]$} & 0 & 0 & 0 & 56.4 \\
Qwen2.5-7B & job-specific direction, best for both (block 15, $c$=0.4) & 28 & 11.8 {\scriptsize $[3.9, 21.2]$} & 10 & 50 & 0 & 56.7 \\
Qwen2.5-7B & job-specific direction, best for flipping (block 10, $c$=0.6) & 25 & 7.0 {\scriptsize $[0.0, 16.4]$} & 0 & 65 & 2 & 48.1 \\
Qwen2.5-7B & random control for the row above (block 10, $c$=0.6) & 61 & 7.5 {\scriptsize $[0.8, 17.7]$} & 0 & 32 & 0 & 56.2 \\
Qwen2.5-7B & general direction, best for both (block 10, $c$=0.4) & 49 & 13.0 {\scriptsize $[3.6, 23.9]$} & 0 & 38 & 0 & 58.9 \\
Qwen2.5-7B & general direction, best for flipping (block 20, $c$=0.4) & 9 & 48.1 {\scriptsize $[28.6, 68.1]$} & 0 & 41 & 2 & 55.6 \\
Qwen2.5-7B & random control for the row above (block 20, $c$=0.4) & 91 & 0.0 {\scriptsize $[0.0, 0.0]$} & 0 & 9 & 0 & 55.5 \\
Llama-3.1-8B & no steering & 100 & 0.0 {\scriptsize $[0.0, 0.0]$} & 0 & 0 & 0 & 56.7 \\
Llama-3.1-8B & job-specific direction, best for both (block 12, $c$=0.4) & 46 & 4.2 {\scriptsize $[0.0, 11.2]$} & 10 & 39 & 0 & 62.1 \\
Llama-3.1-8B & random control for the row above (block 12, $c$=0.4) & 64 & 9.4 {\scriptsize $[0.0, 21.0]$} & 13 & 13 & 0 & 61.1 \\
Llama-3.1-8B & general direction, best for both (block 17, $c$=0.6) & 26 & 37.7 {\scriptsize $[22.4, 54.7]$} & 8 & 28 & 0 & 61.9 \\
Llama-3.1-8B & random control for the row above (block 17, $c$=0.6) & 89 & 0.0 {\scriptsize $[0.0, 0.0]$} & 1 & 7 & 3 & 55.2 \\
\bottomrule
\end{tabular}